\documentclass{article}

\usepackage[numbers]{natbib}

\usepackage[final]{nips_2016}

\usepackage[utf8]{inputenc} 
\usepackage[T1]{fontenc}    
\usepackage{hyperref}       
\usepackage{url}            
\usepackage{booktabs}       
\usepackage{amsfonts}       
\usepackage{nicefrac}       
\usepackage{microtype}      

\usepackage{float}
\usepackage{graphics}
\usepackage{graphicx}
\usepackage[font=small,labelfont=bf]{caption}
\usepackage{subcaption}
\usepackage{enumerate}
\usepackage{amsmath}
\usepackage[ruled,vlined]{algorithm2e}

\usepackage{todonotes}

\title{SPASS: Scientific Prominence Active Search System with Deep Image Captioning Network}

\author{
  Dicong Qiu\thanks{This work was carried out at NASA's Jet Propulsion Laboratory, California Institute of Technology, Pasadena, CA, 91109, U.S.A. Work in progress.}\\
  Robotics Institute\\
  Carnegie Mellon University\\
  Pittsburgh, PA 15213 \\
  \texttt{dq@cs.cmu.edu}
}

\begin{document}

\maketitle

\begin{abstract}
    Planetary exploration missions with Mars rovers are complicated, which generally require elaborated task planning by human experts, from the path to take to the images to capture. NASA has been using this process to acquire over 22 million 
    images from the planet Mars. In order to improve the degree of automation and thus efficiency in this process, we propose a system for planetary rovers to actively search for prominence of prespecified scientific features in captured images. Scientists can prespecify such search tasks in natural language and upload them to a rover, on which the deployed system constantly captions captured images with a deep image captioning network and compare the auto-generated captions to the prespecified search tasks by certain metrics so as to prioritize those images for transmission. As a beneficial side effect, the proposed system can also be deployed to ground-based planetary data systems as a content-based search engine.
\end{abstract}

\section{Introduction}\label{sec:introduction}


NASA has been operating multiple rovers on Mars, including Spirit (MER-A), Opportunity (MER-B) and Curiosity (MSL). In these missions, NASA has acquired more than 22 million 
images from the planet Mars, which were usually generated through the process \cite{gulick2001autonomous} that human experts planned the path to take and the views to shoot photos beforehand, after which photos were taken and transferred sequentially in accord to the prespecified plan. Recent autonomous navigation \cite{carsten2007global} and science targeting \cite{estlin2012aegis} algorithms have been developed for rovers, but they still require human experts to specify higher-level task plans (command sequence).

As more planetary rover missions are put on the agenda and much more imagery data will therefore be generated, the above process seemingly falls short of efficiency: (1) the imagery data are captured in accord to the prespecified plan and will be transmitted sequentially with limited filtering or prioritization through the downlink constrained by limited communications windows, blackouts and narrow bandwidth \cite{gulick2001autonomous, castano2003rover}; (2) the prespecified plans are elaborated by experts who will have to forecast the outcome of their plans, which may not be optimal due to the lack of real-time information and decision making; (3) rovers rely on prespecified command sequence to operate, which implies their travel distance per Martian day (sol) is limited \cite{carsten2007global}, since experts can come up with new command sequence only until new imagery data become available for inspection and more information is therefore gained; (4) The images need manual review and filtering by scientists after transferred.


In order to overcome these disadvantages of the current process, we propose the Scientific Prominence Active Search System (SPASS), which can be deployed on future planetary rovers and actively search for desirable scientific features on the other planets according to search tasks uploaded by scientists described in natural language. SPASS employs an image captioning network architecture with attention mechanism \cite{xu2015show} to provide concise verbal description (caption) for captured images on a rover. And the bilingual evaluation understudy (BLEU) metric \cite{papineni2002bleu} is used to evaluate the similarity between the uploaded search tasks and the generated captions, so as to prioritize the images such that the ones with higher similarity scores will be transmitted first. Moreover, this system can also be deployed as a multi-instance content-based search engine on ground-based planetary data systems, such as NASA's Planetary Data System (PDS) Imaging Atlas\footnote{\url{https://pds-imaging.jpl.nasa.gov}}.

After properly deployed on a rover, SPASS is expected to (1) improve data acquisition efficiency by automating the prioritization process onboard to select high-priority images to transfer in the limited downlink, (2) allow rovers to travel longer distance and explore larger area per sol in collaboration with future autonomous task and path planning algorithms, when manual task and path planning becomes less necessary, (3) reduce human labor on downloaded data reviewing and post-processing, since the images received have been automatically and selectively filtered once onboard.

\section{Related Work}\label{sec:related_work}

People have been using automated systems to analyze planetary surface features for decades. Recent advance in machine learning has also been introduced into aerospace industry for autonomous analysis of surface features from images captured by rovers.

As an image classification system, Deep Mars \cite{wagstaff2018deep} takes advantage of convolutional neural networks (CNNs) to classify images from Mars rovers for engineering-focused objects (such as rover wheels and drill holes) with MSLNet, and orbital images for region types (such craters and dunes) with HiRISENet. However, both MSLNet and HiRISENet were constructed based on AlexNet \cite{krizhevsky2012imagenet}, which can recognize only one object in an image. Pixel-wise segmentation and classification approaches have also been explored by researchers. TextureCam \cite{thompson2012smart} leverages random forest to detect and classify rocks onboard. The Soil Property and Object Classification (SPOC) \cite{rothrock2016spoc} model approaches pixel-wise Mars terrain segmentation by utilizing a fully-convolutional neural network (FCNN). Although these methods are capable of detecting and classifying multiple instances (objects) in an image, they fall short of understanding the relationships among different instances.

End-to-end image captioning is intriguing because it is able to not only detect and classify multiple instances in an image, but also recognize the relationships among them. Vinyals et al. proposed a model \cite{vinyals2015show} that encodes an image using a Deep CNN and generates a complete sentence in natural language to describe the image using a recurrent neural network (RNN) with long short-term memory (LSTM). Xu et al. then proposed to improve the captioning performance and model interpretability by adding a visual attention mechanism between the image encoder and the LSTM network \cite{xu2015show}, based on which the image captioning component in our proposed work was constructed.

\section{Method}\label{sec:method}

SPASS can be deployed on rovers and ground-based data systems, both of which use an image captioning network architecture with attention mechanism to caption captured images. A text similarity metric, the BLEU metric, is then applied to prioritize the images according to the uploaded search tasks. In addition, a data aggregation and model retraining pipeline is developed for ground-based systems, in order to improve the image captioning model weights over time.

\subsection{Image Captioning with Visual Attention}\label{sec:method_image_captioning}

\begin{figure}[H]
    \centering
    \includegraphics[width=0.72\linewidth]{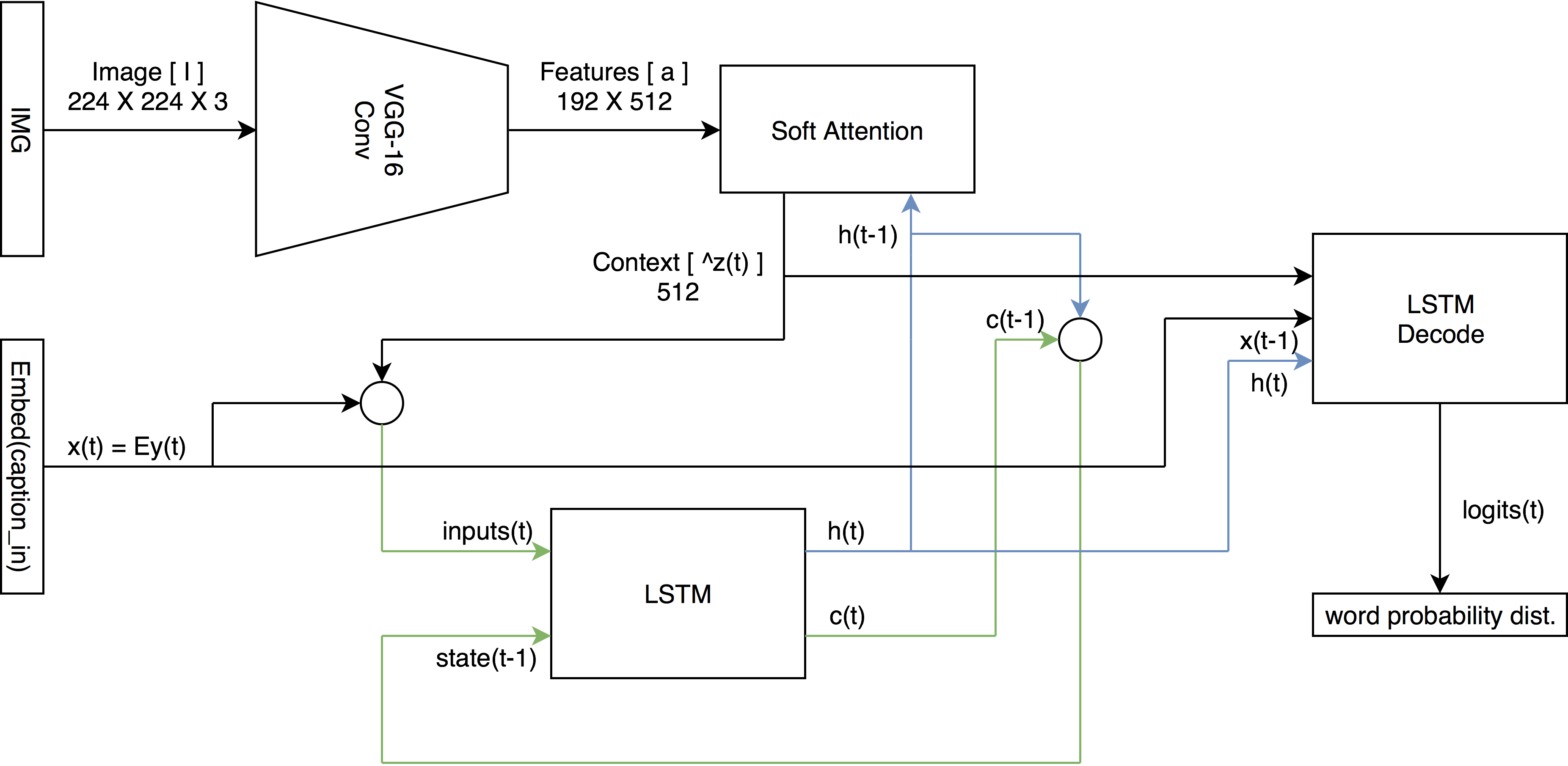}
    \caption{Architecture of the deep image captioning model with visual attention mechanism.}
    \label{fig:architecture_image_captioning}
\end{figure}

The image captioning model in SPASS consists of an image encoder (feature extractor), an attention mechanism and a LSTM network \cite{hochreiter1997long}, as shown in figure \ref{fig:architecture_image_captioning}, which is constructed mostly referring to the model architecture proposed by Xu et al. \cite{xu2015show}. The model takes a single raw image and generates a caption $ y $ encoded as a sequence of 1-of-$K$ encoded words.

\[
y = \left\{ \mathbf{y}_{1},\cdots,\mathbf{y}_{C} \right\}, \mathbf{y}_{i} \in \mathbb{R}^{K}
\]

where $ K $ is size of the vocabulary and $ C $ is the length of the caption.

The image encoder extract convolutional features from a raw input image, and can be the convolutional layers of any convolutional neural network, such as VGGNet \cite{simonyan2014very}, GoogLeNet \cite{szegedy2015going} or ResNet \cite{he2016deep}. Among these alternatives, we used the VGG-19 architecture, which takes an input image resized to $ 224 \times 224 \times 3 $ and generates a $ 14 \times 14 \times 512 $ feature map. The extracted feature map is then flatten to $ L \times D = (14 \times 14) \times 512 = 192 \times 512 $.

The flatten features $ \mathbf{a} = \left\{ \mathbf{a}_{1},\cdots,\mathbf{a}_{L} \right\} $ joins the hidden state vector $ \mathbf{h}_{t-1} $ from the last step of the LSTM network (which will be discussed in the later) to form a joint vector that goes through a fully connected attention network, which generates attention weights $ \alpha = \left\{ \alpha_{1},\cdots,\alpha_{L} \right\} $ for each of the $ L = 192 $ features. The attention weights are assigned to features with the soft attention mechanism proposed by Bahdanau et al. \cite{bahdanau2014neural}, outputting a context vector $ \hat{\mathbf{z}}_{t} $ at time step $ t $.

\[
\begin{split}
    & e_{t,i} = f_{\text{att}} \left( \mathbf{a}_{i}, \mathbf{h}_{t-1} \right) \\
    & \alpha_{t,i} = \frac{ \text{exp} \left( e_{t,i} \right) }{ \sum_{k=1}^{L} \text{exp} \left( e_{t,k} \right) } \\
    & \hat{\mathbf{z}}_{t} = \phi \left( \left\{ \alpha_{t,i} \right\}, \left\{ \mathbf{a}_{i} \right\} \right) = \sum_{i=1}^{L} \alpha_{t,i} \mathbf{a}_{i}
\end{split}
\]

where $ \mathbf{h}_{t-1} $ is the LSTM hidden state, $ f_{\text{att}} $ represents the attention weights generation network and $ \phi $ is the soft attention weights assignment mechanism.

\begin{figure}[H]
    \centering
    \includegraphics[width=0.45\linewidth]{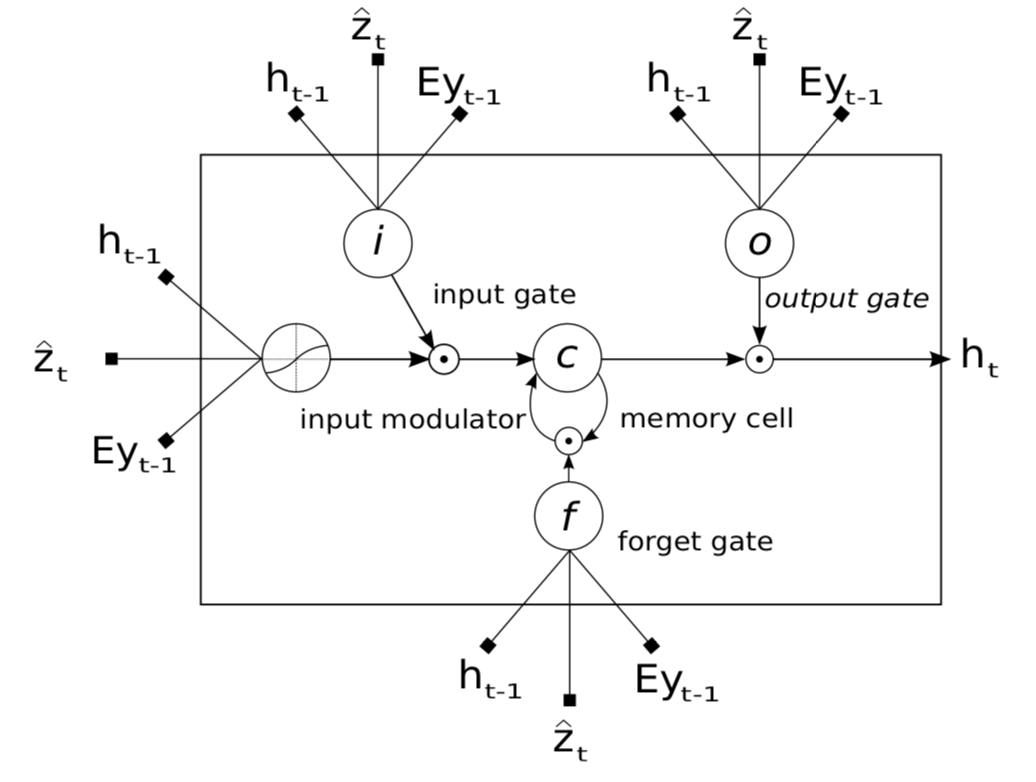}
    \caption{The long short-term memory (LSTM) network architecture \cite{xu2015show}.}
    \label{fig:LSTM}
\end{figure}

A following LSTM network, which was constructed closely following the one in \cite{zaremba2014recurrent}, takes the previous hidden state $ \mathbf{h}_{t-1} $, the current context vector $ \hat{\mathbf{z}}_{t} $ and the previous word embedding $ \mathbf{E} \mathbf{y}_{t-1} $, and generates the next hidden state $ \mathbf{h}_{t} $.

\[
\begin{split}
    \mathbf{i}_{t} &= \sigma \left( \mathbf{W}_{i} \mathbf{E} \mathbf{y}_{t-1} + \mathbf{U}_{i} \mathbf{h}_{t-1} + \mathbf{Z}_{i} \hat{\mathbf{z}}_{t} + \mathbf{b}_{i} \right) \\
    \mathbf{f}_{t} &= \sigma \left( \mathbf{W}_{f} \mathbf{E} \mathbf{y}_{t-1} + \mathbf{U}_{f} \mathbf{h}_{t-1} + \mathbf{Z}_{f} \hat{\mathbf{z}}_{t} + \mathbf{b}_{f} \right) \\
    \mathbf{c}_{t} &= \mathbf{f}_{t} \mathbf{c}_{t-1} + \mathbf{i}_{t} \tanh \left( \mathbf{W}_{c} \mathbf{E} \mathbf{y}_{t-1} + \mathbf{U}_{c} \mathbf{h}_{t-1} + \mathbf{Z}_{c} \hat{\mathbf{z}}_{t} + \mathbf{b}_{c} \right) \\
    \mathbf{o}_{t} &= \sigma \left( \mathbf{W}_{o} \mathbf{E} \mathbf{y}_{t-1} + \mathbf{U}_{o} \mathbf{h}_{t-1} + \mathbf{Z}_{o} + \hat{\mathbf{z}}_{t} + \mathbf{b}_{o} \right) \\
    \mathbf{h}_{t} &= \mathbf{o}_{t} \tanh \left( \mathbf{c}_{t} \right)
\end{split}
\]

where $ \mathbf{i}_{t} $, $ \mathbf{f}_{t} $, $ \mathbf{c}_{t} $, $ \mathbf{o}_{t} $ and $ \mathbf{h}_{t} $ are respectively the input, forget, memory, output and hidden states of the LSTM at time step $ t $; $ \mathbf{W}_{\bullet} $, $ \mathbf{U}_{\bullet} $, $ \mathbf{Z}_{\bullet} $ and $ \mathbf{b}_{\bullet} $ are learned parameters; $ \mathbf{E} \in \mathbb{R}^{m \times K} $ is an embedding matrix. $ m $ denotes the embedding dimensionality. With the generated sentence encoded as a sequence of $1$-of-$K$ encoded words $ y = \left\{ \mathbf{y}_{1}, \cdots, \mathbf{y}_{C} \right\} $, each word in the sentence is selected from the a word probability distribution $ \mathbf{y}_{t} $ generated using a deep output layer by decoding the hidden state $ \mathbf{h}_{t} $.

\[
p \left( \mathbf{y}_{t} \middle| \mathbf{a}, \mathbf{y}_{1}^{t-1}
\right) \propto \text{exp} \left( \mathbf{L}_{o} \left( \mathbf{E} \mathbf{y}_{t-1} + \mathbf{L}_{h} \mathbf{h}_{t-1} + \mathbf{L}_{z} \hat{\mathbf{z}}_{t} \right) \right)
\]

where $ \mathbf{L}_{o} \in \mathbb{R}^{K \times m} $, $ \mathbf{L}_{h} \in \mathbb{R}^{m \times n} $ and $ \mathbf{L}_{z} \in \mathbb{R}^{m \times D} $ are learned parameters. $ n $ and $ D = 512 $ denote the LSTM and feature dimensionality, respectively. $ K $ is the size of the vocabulary and $ C $ is the length of the caption.



\subsection{Similarity Evaluation between Captions and Search Tasks}\label{sec:method_text_similarity_evaluation}

Evaluation of the text similarity between the auto-generated image captions and the uploaded search tasks leads to similarity scores, according to which the captured images can be prioritized. Text similarity evaluation for natural language applies similar principles of most machine translation (MT) evaluation techniques. A variety of such techniques has been developed over time \cite{reeder2001additional}, among which the BLEU metric \cite{papineni2002bleu} is used in SPASS to provide similarity scores between image captions and search tasks. BLEU metric evaluates a similarity score between multiple MT references and candidates. In SPASS, uploaded search tasks are considered as MT references and the auto-generated caption for a captured image is considered as a MT candidate. A higher BLEU score indicates a higher relevance between the search tasks and the captured image, and hence the image will gain a higher priority for transmission.

An improvement BLEU metric made from standard $n$-gram precision measure is that it uses a modified $n$-gram precision measure. It first computes the $n$-gram matches, then adds the clipped $n$-gram counts for all candidate sentences, and finally divides the sum by the number of candidate $n$-grams to compute a modified precision score $ p_{n} $.

\[
p_{n}\left( c, R \right) = \frac{ \sum_{ u \in G_{n}\left( c \right) } \min\left[ \mathbb{I}\left( u, R \right) N_{c}\left( u \right), \max_{ r \in R } N_{r}\left( u \right) \right] }{ \sum_{ u' \in G_{n} \left( c \right) } N_{c}\left( u' \right) }
\]

where $ R $ is the collection of the uploaded search tasks, $ c $ is the auto-generated caption for the captured image, $ G_{n}\left( c \right) $ is the collection of $n$-grams in caption $ c $, $ N_{t}\left( u \right) $ is the count of the $n$-gram $ u $ for text block $ t $, and $ \mathbb{I}\left( u, T \right) $ is an indicator function which becomes $ 1 $ if any text block $ t $ in text block collection $ T $ contains the $n$-gram $ u $ otherwise $ 0 $.

Such an evaluation metric formulation, however, tends to favor shorter auto-generated captions (MT candidates). To overcome this drawback, a multiplicative brevity penalty $ \eta $ is introduced to discourage auto-generated captions that are too short, which convey limited information. On the other hand, auto-generated captions longer than uploaded tasks will receive no extra penalty since they have been penalized by the $n$-gram precision score.

\[
\eta\left( c, r \right) = \left\{\begin{matrix}
    1 & \text{if } L(c) > L(r) \\
    \exp\left( 1 - \frac{L(r)}{L(c)} \right) & \text{if } L(c) \le L(r)
\end{matrix}\right.
\]

with $ c $ being an auto-generated caption, $ r $ an uploaded search task, and $ L\left( t \right) $ the length of text block $ t $.

To put together, captured images will be automatically captioned and prioritized for transmission in accord to the their similarity scores. The similarity score $ s $ for an auto-generated caption under a search task set $ R $ is evaluated considering the $n$-gram precision and the caption brevity.

\[
s_{N}\left(c, R\right) = \eta\left( c, \arg\max_{r \in R} L(r) \right) \exp\left( \sum_{n=1}^{N} w_{n} \log p_{n}\left(c, R\right) \right)
\]

Its logarithm version is sometimes even more useful in practical scenarios.

\[
\log s_{N}\left(c, R\right) = \min\left[ 1 - \frac{ \max_{r \in R} L(r) }{ L(c) }, 0 \right] + \sum_{n=1}^{N} w_{n} \log p_{n}\left(c, R\right)
\]

where $ N $ is the maximum length up to which $n$-grams are used and $w_{n}$ is the importance weight assigned to the $n$-gram precision measure. In our system, $ N = 4 $ and the weights $ \left\{ w_{n} \right\}_{n=1}^{4} = \left\{ 0.8, 0.15, 0.045, 0.005 \right\} $ are used.

\subsection{Data Aggregation and Model Retraining Pipeline}\label{sec:method_pipeline}

The quantity and variance of data are vital to a machine learning system. To the best of our knowledge, there is no Martian rover dataset that is accessible to the public and applicable to the training of image captioning models. To address the problem of insufficient training data, a data aggregation and model retraining pipeline was proposed, along with its deployment onto an online web-based tool, which will be further discussed in the following section \ref{sec:method_web_tool}.

The pipeline starts with an annotated initial dataset. It is used to train the initial image captioning model, which can be deployed to rovers or ground-based auto-captioning system to caption images without annotations. These captions will then be uploaded to an online annotation tool for multiple expert reviewers to review, correct or refine. Reviewed image caption pairs will be augmented to the dataset, which, after sufficient new data are augmented to, will be used to retrain the model to improve its performance. The following is a concrete procedure.

\begin{algorithm}[H]
\SetAlgoLined
\KwResult{Refined model weights sequence $ \mathcal{W} $ and aggregated dataset $ \mathcal{D} = \{ \mathcal{I}, \mathcal{A} \} $.}
 Input model $ M $, initial dataset $ \mathcal{D}_{0} = \{ \mathcal{I}_{0}, \mathcal{A}_{0} \} $ and aggregation rate $ \delta > 0 $\;
 Randomly initialize the initial model weights $ W_{0} $\;
 Initialize aggregated image dataset $ \mathcal{I} \leftarrow \mathcal{I}_{0} $, annotation dataset $ \mathcal{A} \leftarrow \mathcal{A}_{0} $ and let $ \mathcal{D} \leftarrow \{ \mathcal{I}, \mathcal{A} \} $\;
 Initialize training dataset cache $ \mathcal{D}_{\text{train}} \leftarrow \{\} $ and model weights sequence $ \mathcal{W} \leftarrow \{ W_{0} \} $\;
 \While{True}{
  \If{$ \left| \mathcal{D}_{\text{train}} \right| \times \left( 1 + \delta \right) \le \left| \mathcal{D} \right| $}{
   Cache training dataset $ \mathcal{D}_{\text{train}} \leftarrow \mathcal{D} $\;
   Fit model with training data so that $ W_{\left| \mathcal{W} \right|} \leftarrow \textit{model\_fit}\left( M, W_{\left| \mathcal{W} \right| - 1}, \mathcal{D}_{\text{train}} \right) $\;
   Update model weights sequence $ \mathcal{W} \leftarrow \mathcal{W} \bigcup \{ W_{\left| \mathcal{W} \right|} \} $\;
  }
  Receive new captured images $ \mathcal{I}_{\text{new}} $\;
  Annotate new images with the latest trained model $ \hat{\mathcal{A}}_{\text{new}} \leftarrow \textit{model\_predict}\left( M, W_{\left| \mathcal{W} \right| - 1}, \mathcal{I}_{\text{new}} \right) $\;
  Open review for $ \hat{\mathcal{A}}_{\text{new}} $ and receive updated annotation set $ \mathcal{A}_{\text{new}} $\;
  Update aggregated datasets $ \mathcal{I} \leftarrow \mathcal{I} \bigcup \mathcal{I}_{\text{new}} $, $ \mathcal{A} \leftarrow \mathcal{A} \bigcup \mathcal{A}_{\text{new}} $ and $ \mathcal{D} \leftarrow \{ \mathcal{I}, \mathcal{A} \} $\;
 }
 \caption{Data Aggregation and Model Retraining Procedure}
\end{algorithm}

In the above algorithm, $ \textit{model\_fit} $ is the model fitting procedure and $ \textit{model\_predict} $ is the model prediction procedure as discussed in section \ref{sec:method_image_captioning}. In practice, the data aggregation rate $ \delta = 0.5 $ in our system. And we also created a dataset as our initial dataset using the online web tool we developed (see: section \ref{sec:method_web_tool}). The dataset has 1322 images captured by the MSL rover along with corresponding expert-annotated captions, which includes primarily Martian geological science data, such as images of different terrains and rocks.

An intuitive illustration of the aforementioned procedure is shown below in figure \ref{fig:data_pipeline}.

\begin{figure}[H]
    \centering
    \includegraphics[width=0.9\linewidth]{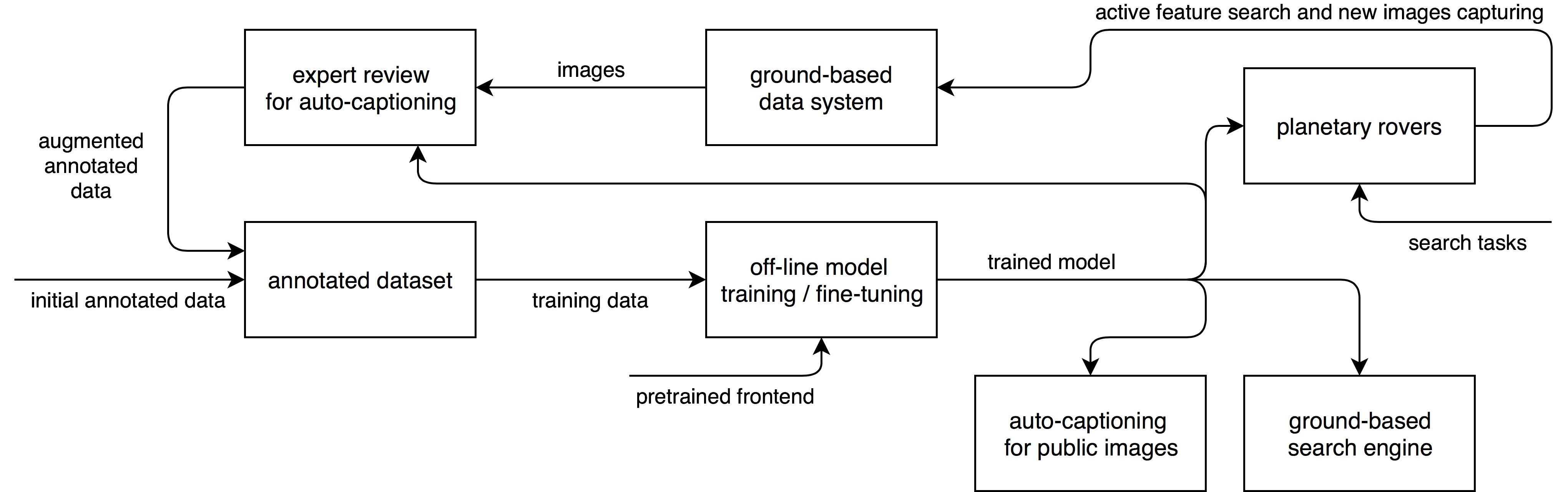}
    \caption{An illustration of the data aggregation and model retraining pipeline.}
    \label{fig:data_pipeline}
\end{figure}


\subsection{Online Web Tool}\label{sec:method_web_tool}

To support multiple users in providing annotation reviews for incoming images and uploading search tasks, an online multi-user web-based tool was developed, on which a ground-based version of SPASS was deployed. This web tool can also serve as a ground-based planetary data system.

\begin{figure}[H]
    \centering
    \begin{subfigure}{0.32\textwidth}
        \centering
        \includegraphics[width=\linewidth]{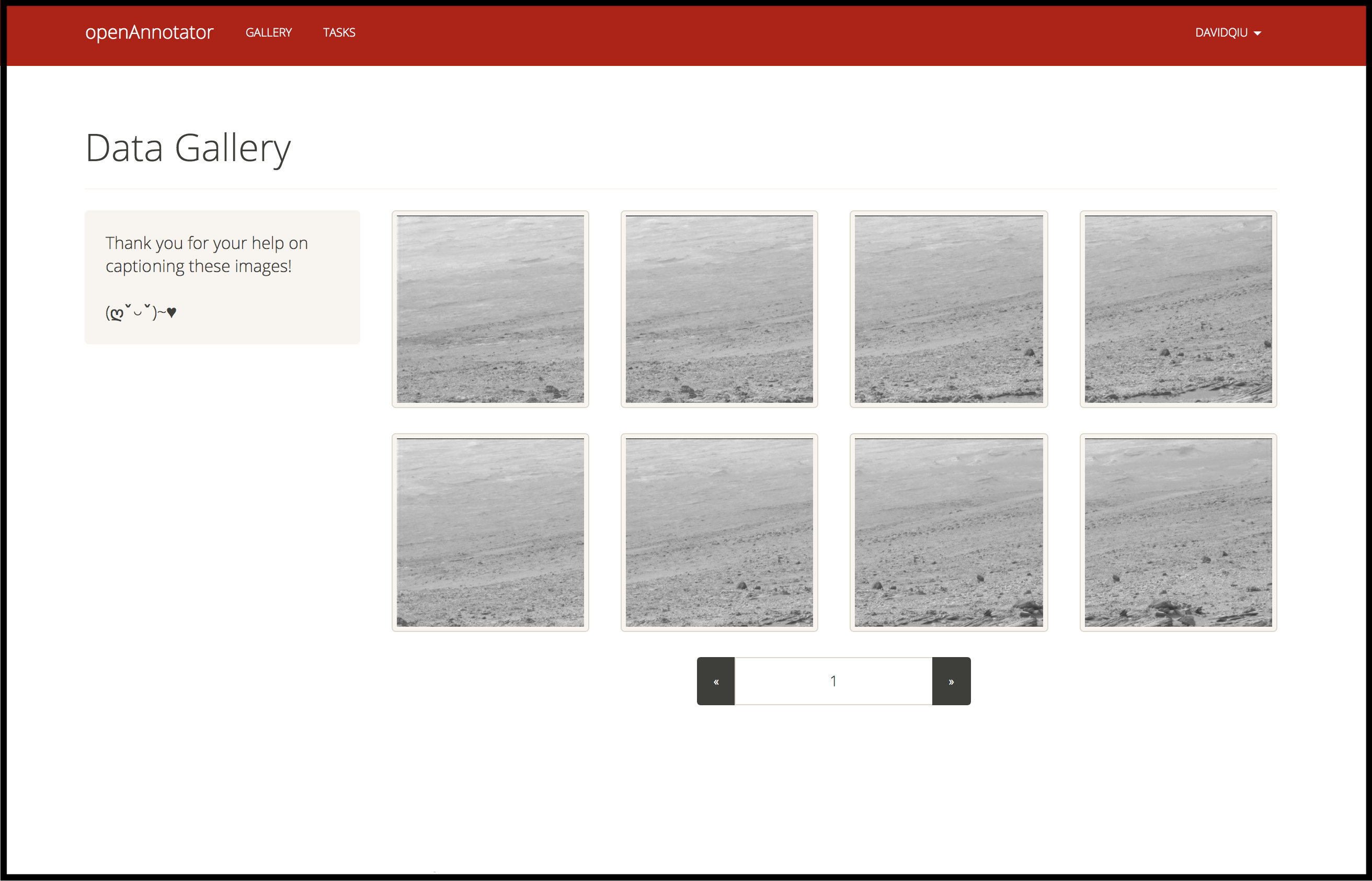}
        \caption{Gallery Page}
        \label{fig:openAnnotatorOnline_UI_Gallery}
    \end{subfigure}
    \begin{subfigure}{0.32\textwidth}
        \centering
        \includegraphics[width=\linewidth]{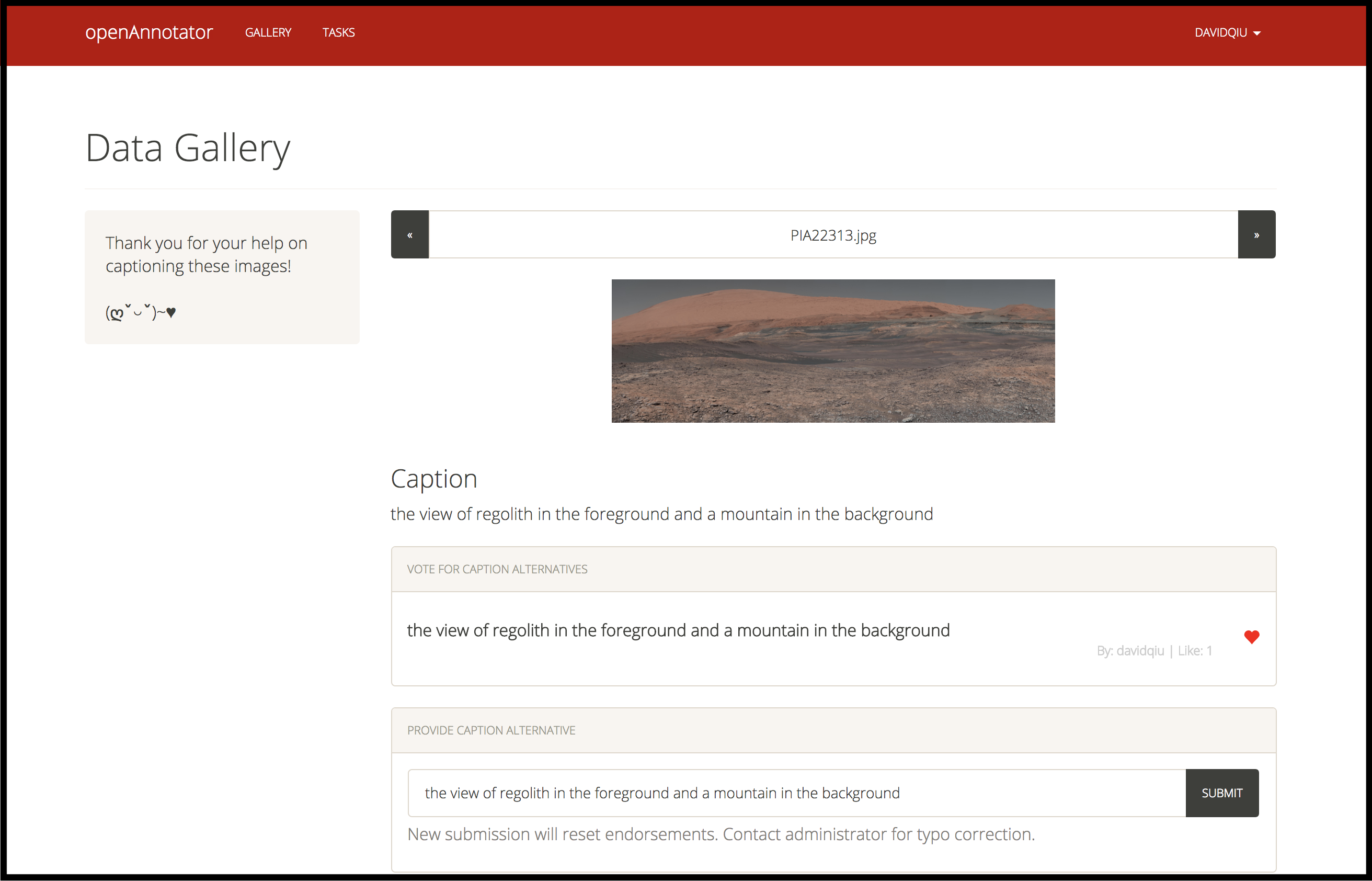}
        \caption{Annotation Page}
        \label{fig:openAnnotatorOnline_UI_Image}
    \end{subfigure}
    \begin{subfigure}{0.32\textwidth}
        \centering
        \includegraphics[width=\linewidth]{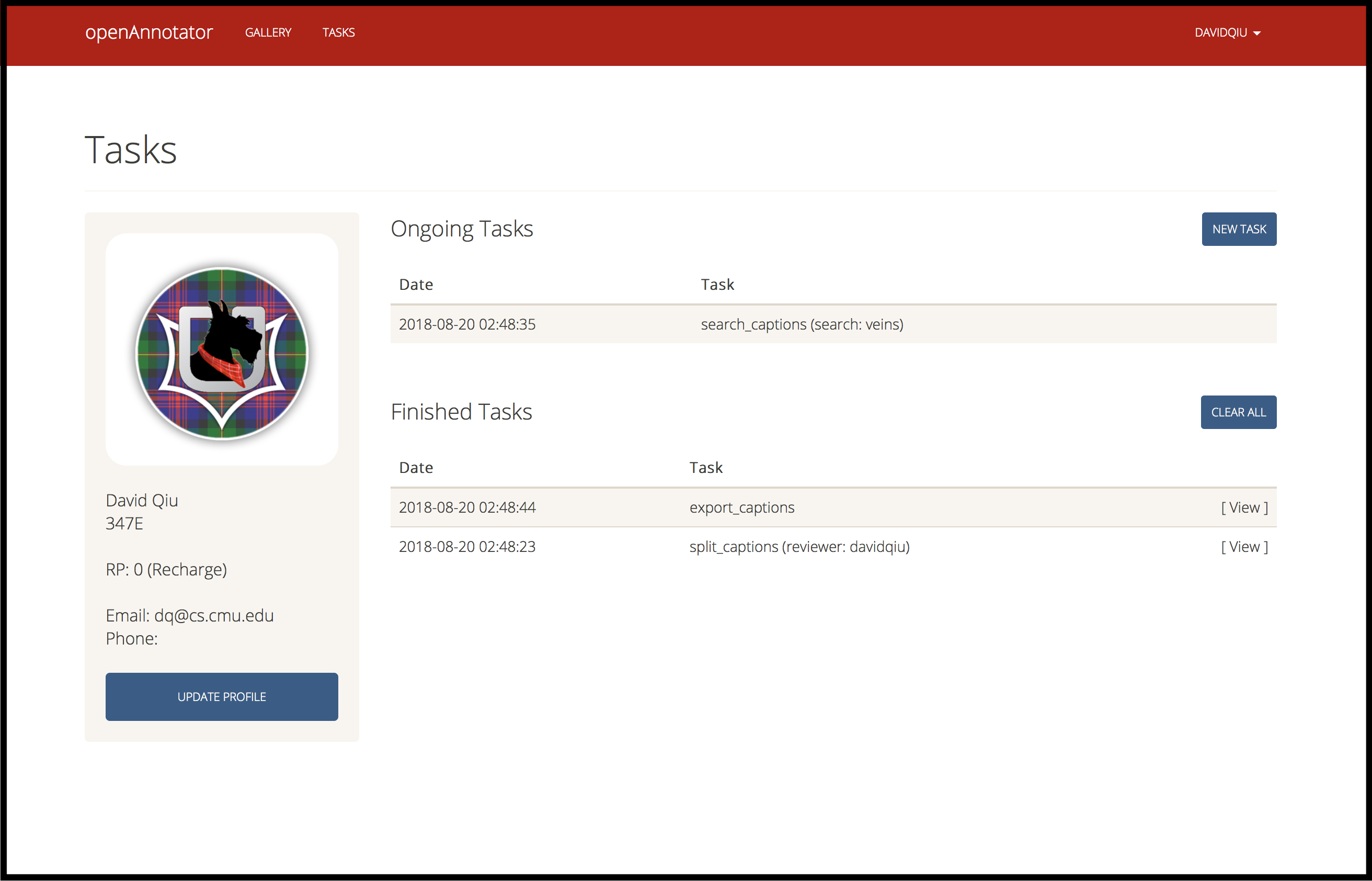}
        \caption{Tasks Page}
        \label{fig:openAnnotatorOnline_UI_Tasks}
    \end{subfigure}
    \caption{The user interface of the web tool we developed.}
    \label{fig:openAnnotatorOnline_UI}
\end{figure}

The web tool allows multiple users to (1) collaborate in annotating images at the same time, (2) provide their own versions of captions, (3) vote for their favorite captions, (4) upload ground-based search tasks to search through the local planetary data, and (5) export the latest training dataset and model weights. Moreover, since the data aggregation and model retraining pipeline (see: section \ref{sec:method_pipeline}) was deployed as well onto the web tool, it will trigger model retraining at specific moments as more data are aggregated and thus improve the model performance over time.

\section{Experiment}\label{sec:experiment}


As mentioned in section \ref{sec:method_pipeline}, the dataset \textit{Mars1322} has been created with 1322 image-caption pairs of science-focus Martian surface features, which was used in this section to train and evaluate the attention image captioning model. In our setup, the maximum caption length $ C_{\max} = 20 $, under which data samples with caption length $ C > C_{\max} $ were removed and $ 1242 $ samples remained. These samples were randomly split into a training dataset and a validation dataset by a ratio of $ 0.90:0.10 $. All images were resized to $ 224 \times 224 $ during data preprocessing.

We used the VGG-19 network convolutional layers as image feature extractor (encoder). In our experiment, the weights of the convolutional layers were pretrained on ImageNet \cite{deng2009imagenet}, and during the image captioning model training procedure the these weights were fixed. In other words, only the weights of the attention layers and the LSTM network were randomly initialized and trained from scratch on the domain relevant data. As for the training procedure, stochastic gradient descent (SGD) with adaptive learning rate, more specifically the Adam algorithm \cite{kingma2014adam}, was used is used. We also applied dropout \cite{srivastava2014dropout} and early stopping on BLEU score to prevent overfitting.

\begin{figure}[H]
    \centering
    \includegraphics[width=0.6\linewidth]{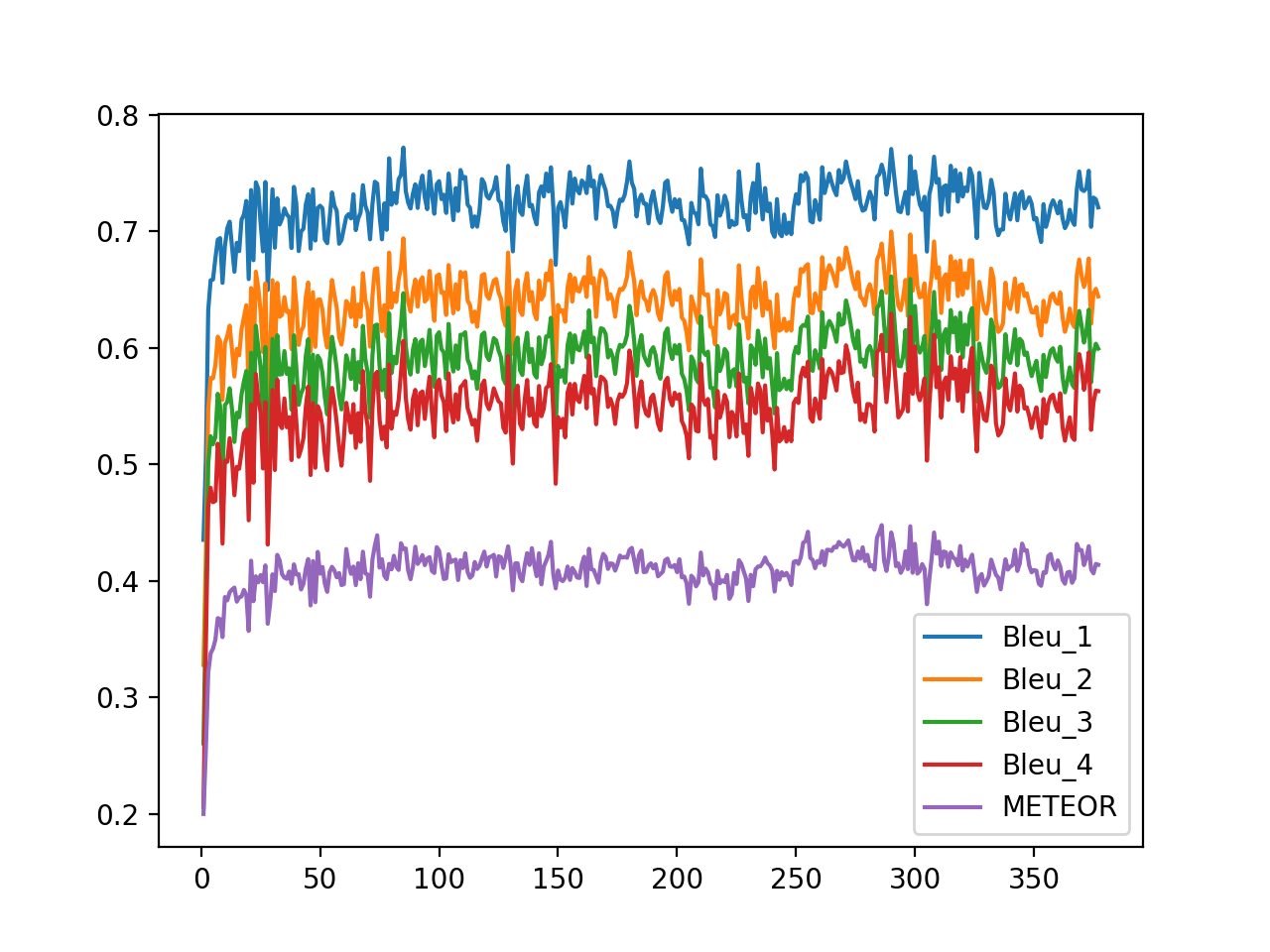}
    \caption{BLEU-1,2,3,4/METEOR scores on validation data along the training procedure.}
    \label{fig:learning_curve}
\end{figure}

In the training procedure, scores evaluated by BLEU-1,2,3,4 and METEOR \cite{banerjee2005meteor} metrics on validation dataset were taken down. Figure \ref{fig:learning_curve} visualizes the update of these scores along the $ 377 $ training epochs.

After its attention layers and LSTM network trained, the image captioning model achieved BLEU and METEOR scores that indicate the trained model works reasonably well in this confined domain, in compare to the scores reported in \cite{xu2015show} where the model confronted more complicated domains as implicitly defined by the \textit{Flickr8k}, \textit{Flickr30k} and \textit{MS COCO} datasets.


\begin{table}[H]
\centering
\begin{tabular}{c|cccc|c}
\cline{2-5}
                                         & \multicolumn{4}{c|}{\textbf{BLEU}}                                                               &                             \\ \hline
\multicolumn{1}{|c|}{Dataset}            & \multicolumn{1}{c|}{BLEU-1} & \multicolumn{1}{c|}{BLEU-2} & \multicolumn{1}{c|}{BLEU-3} & BLEU-4 & \multicolumn{1}{c|}{METEOR} \\ \hline
\multicolumn{1}{|c|}{\textit{Mars1322}}  & 0.7717                      & 0.6997                      & 0.6611                      & 0.6293 & \multicolumn{1}{c|}{0.4478} \\ \hline
\end{tabular}
\caption{Model evaluation by BLEU-1,2,3,4/METEOR metrics.}
\label{tab:metrics}
\end{table}




\section{Conclusion}\label{sec:conclusion}

We propose a content-based search system, SPASS, for future Mars (planetary) rovers to improve data transmission efficiency by actively searching for more scientifically valuable imagery data. Our approaches for automatic image caption generation and priority (relevant to search tasks) evaluation were presented. And a data aggregation and model retraining pipeline was developed and deployed onto an online web tool. In the experiment section, we also show that our approach achieved a decent performance in the restricted domain of Martian surface feature analysis as evaluated by BLEU and METEOR metrics.

\section*{Acknowledgement}\label{sec:acknowledgement}

The work described in this paper was carried out at NASA's Jet Propulsion Laboratory, California Institute of Technology. The author would also like to thank Masahiro Ono and Brandon Rothrock for advising, Tanvir Islam, Annie K. Didier, Bhavin T. Shah and Chris A. Mattmann for supporting this work, and Adrian Stoica for providing workspace.

\bibliographystyle{unsrt}
\bibliography{refs}

\end{document}